\documentclass[10pt,conference]{IEEEtran}
\usepackage[utf8]{inputenc} 
\usepackage{cite}
\usepackage[numbers]{natbib}
\usepackage[numbers]{natbib}
\usepackage{makecell}
\expandafter\let\csname equation*\endcsname\relax
\expandafter\let\csname endequation*\endcsname\relax
\usepackage{amsmath,amssymb,amsfonts}
\usepackage{algorithmic}
\usepackage{graphicx}
\usepackage{textcomp}
\usepackage{tabularx}
\usepackage{xcolor}
\usepackage{multirow}
\usepackage{physics}
\usepackage{braket}
\usepackage{tikz}
\usetikzlibrary{quantikz2}
\usepackage{booktabs}
\usepackage{float}
\usepackage{subcaption}
\usepackage{braket}
\usepackage{tikz}
\usepackage{placeins}

\def\BibTeX{{\rm B\kern-.05em{\sc i\kern-.025em b}\kern-.08em
    T\kern-.1667em\lower.7ex\hbox{E}\kern-.125emX}}
\begin{document}

\title{Empirical Characterization of Learning Geometry in Hybrid Quantum Forecasting Models}

\author{
\IEEEauthorblockN{
Sandra Leticia Juárez-Osorio\IEEEauthorrefmark{1},
Jorge I. Hernandez-Martinez\IEEEauthorrefmark{1},
Jesus Ivan Ruiz-Martinez \IEEEauthorrefmark{1},
Andres Mendez-Vazquez\IEEEauthorrefmark{1},\\
Eduardo Rodriguez-Tello\IEEEauthorrefmark{2}
}

\IEEEauthorblockA{\IEEEauthorrefmark{1}
Department of Computer Science, CINVESTAV Guadalajara, Mexico
}

\IEEEauthorblockA{\IEEEauthorrefmark{2}
Cinvestav Unidad Tamaulipas, Mexico
}

\IEEEauthorblockA{
Email:\{sandra.juarez, jivan.hernandez, jesus.ruiz, andres.mendez, ertello\}@cinvestav.mx
}
}
\maketitle

\begin{abstract}
We characterize the learning dynamics of a compact hybrid quantum forecasting model through comparison with a structurally aligned classical baseline. Using stationary harmonic-mixture and nonstationary chirp benchmarks with controlled spectral complexity and data availability, we analyze empirical Neural Tangent Kernel dynamics through kernel-target alignment, kernel drift, spectral concentration, and training loss. The classical model exhibits stronger early target alignment, whereas the hybrid model generally develops a less concentrated kernel spectrum and smaller kernel drift. Despite these distinct optimization geometries, both architectures attain similar held-out performance across the evaluated regimes. Notably, the hybrid model uses 125 trainable parameters compared with 281 for the classical baseline and reaches its validation-selected checkpoint earlier in 15 of 18 frequency conditions. A Fourier-augmented classical baseline does not reproduce the observed training behavior, while a controlled re-uploading ablation shows that repeated encoding systematically modifies both optimization and kernel geometry. These results demonstrate that comparable generalization can emerge from substantially different learning trajectories and that individual NTK diagnostics do not provide monotonic predictors of validation convergence. Rather than claiming a general quantum advantage, the study identifies architecture-dependent learning behavior that is masked by endpoint accuracy alone.
\end{abstract}


\section{Introduction}

Neural networks are known to exhibit structured learning biases during optimization. One of the most widely studied phenomena is spectral bias, where lower-frequency or smoother components of a target function are typically learned earlier than higher-frequency components \citep{rahaman}. This behavior has been analyzed theoretically through the Neural Tangent Kernel (NTK) framework, where training dynamics can be interpreted through the eigenspectrum of an induced kernel \citep{ntk}. In this view, components associated with larger eigenvalues are learned more rapidly, while smaller-eigenvalue directions converge more slowly. As a result, kernel geometry provides a principled lens for understanding optimization speed, inductive bias, and generalization behavior.

These ideas raise a natural question in quantum machine learning: Do hybrid quantum models induce learning dynamics that differ in a meaningful way from comparable classical architectures? Variational quantum circuits (VQCs) have been proposed as trainable nonlinear feature maps defined through quantum state evolution and parametrized unitary transformations, potentially capturing correlations differently from standard neural networks \citep{variational,centric}. However, many comparisons between quantum and classical models focus primarily on final predictive accuracy, often leaving the underlying optimization geometry less explored. Understanding learning dynamics is particularly relevant in forecasting tasks involving oscillatory or spectrally rich signals. In such problems, differences in learning geometry may become visible not only through final error, but also through how quickly models learn distinct frequency components, how stable their kernels remain during training, and how concentrated their spectral representations become.

In this work, we investigate these questions through a controlled empirical study comparing a compact hybrid quantum model against a classical baseline with matched structure on synthetic multivariate forecasting tasks with tunable spectral complexity. We consider both stationary harmonic-mixture signals and nonstationary chirp signals, while varying training set size to probe data limited regimes.

Rather than focusing exclusively on final predictive performance, we analyze optimization behavior using empirical NTK diagnostics computed during training. Specifically, we track kernel-target alignment, kernel drift relative to initialization, top-1 spectral mass, and learning efficiency through the area under the learning curve (AULC). Validation loss is monitored throughout training to identify when each model reaches its best validation-selected solution, which is subsequently evaluated on test set. 

Our main contributions are summarized as follows:
\begin{itemize}
    \item We introduce an empirical NTK framework for comparing hybrid quantum and classical learning dynamics under controlled spectral complexity and data availability.

    \item We identify persistent architecture-dependent differences in kernel alignment, drift, and spectral concentration across stationary and nonstationary forecasting benchmarks.

    \item We show that the 125-parameter hybrid model attains held-out performance within a few percent of a 281-parameter classical baseline and reaches its validation-selected checkpoint earlier in most evaluated conditions.

    \item Through Fourier-feature and re-uploading studies, we show that explicit periodic features alone do not reproduce the observed training dynamics and that individual NTK diagnostics are not monotonic predictors of convergence.
\end{itemize}

The remainder of this paper is organized as follows: Section II reviews the relevant background, Section III describes the methodology and experimental setup, Section IV presents the results, and Section V concludes the paper.
\section{Background and Related Work}
\label{background}

The optimization dynamics of neural networks are strongly shaped by the
frequency structure of the target function, often
learning simpler functions before more complex ones. Rahaman
et al.~\citep{rahaman} showed that networks trained with gradient-based
methods preferentially fit low-frequency components of a target before
capturing higher-frequency details, a phenomenon known as
\textit{spectral bias}. Their Fourier-based analysis further showed that low-frequency components are more robust to parameter perturbations, highlighting an asymmetry in how neural networks learn and represent different spectral components.

The Neural Tangent Kernel (NTK) framework provides a principled view of
these learning dynamics~\citep{ntk}. In the infinite-width limit, the
evolution of a neural network under gradient descent can be described as
kernel gradient flow in function space, governed by a deterministic
kernel that remains effectively constant during training. Decomposing
the dynamics along the kernel eigenfunctions gives
\begin{equation}
    f_t =
    \sum_i
    \left(1-e^{-\lambda_i t}\right)
    \langle y,\phi_i\rangle\phi_i,
\end{equation}
where $\lambda_i$ and $\phi_i$ denote the eigenvalues and eigenfunctions
of the kernel. Components associated with larger eigenvalues are
therefore learned more rapidly, whereas directions associated with
smaller eigenvalues converge more slowly. The NTK eigenspectrum thus
provides a direct connection between kernel geometry and optimization
dynamics.

The spectral properties of the NTK also characterize its inductive bias.
Bietti and Mairal~\citep{bietti2019inductive} analyzed the reproducing
kernel Hilbert space induced by the NTK through spherical harmonics and
showed that eigenvalues decay with increasingly oscillatory components.
Consequently, smoother functions are preferentially represented, while
more complex components receive progressively smaller spectral weights.
Their analysis also identifies a trade-off between smoothness and
approximation capacity, since slower spectral decay allows richer
function classes at the cost of reduced regularity.

These theoretical predictions are supported by empirical studies.
Cao et al.~\citep{cao} tracked the optimization of functions constructed
from spherical harmonics and observed that lower-degree components are
learned earlier than higher-degree components, consistently with the
spectral dynamics predicted by the NTK framework. Together, these
results motivate the use of kernel eigenspectra and their evolution
during training as empirical diagnostics of learning dynamics and
inductive bias.

\subsection{Quantum Computing and Variational Quantum Circuits}

Quantum computing exploits properties such as superposition and
entanglement to process information through quantum state evolution
\citep{nielsen,40years}. Beyond quantum algorithms for tasks such as
factoring and search~\citep{Shor,Grover}, these properties have motivated
the development of quantum machine learning models that use quantum
circuits as trainable computational components, particularly in hybrid
quantum--classical settings.

Variational Quantum Circuits (VQCs), also called parametrized quantum circuits, are among the most widely studied models for near-term quantum machine learning~\citep{variational,centric}. A VQC combines classical data encoding, layers of parametrized unitary gates, and observable measurements. Its trainable circuit parameters are typically optimized by a classical optimizer, forming a hybrid learning loop.
Because VQCs can be constructed with relatively shallow circuits, they are commonly investigated in the context of noisy intermediate-scale quantum (NISQ) devices \cite{variational}.

Classical data must first be mapped to a quantum state through an encoding operation. Common strategies include amplitude and angle encoding~\citep{libro_supervised}. In angle encoding, classical input values parameterize single-qubit rotations, introducing data-dependent amplitudes or relative phases according to the rotation axis and state preparation. The encoded state is subsequently transformed by trainable quantum gates and entangling
operations, and observable measurements produce classical outputs that can be used as learned features or predictions.

The representational properties of a VQC depend on its encoding strategy, circuit depth, parametrization, and entangling topology~\citep{centric,expressibility}. Data re-uploading extends this construction by repeatedly encoding the input at multiple circuit layers, rather than introducing it only once. This repeated encoding can
increase the class of input-dependent functions represented by the circuit \cite{yo}. In this work, we employ rotation encoding with data re-uploading; the specific architecture is defined in Section~III-B.

Here, VQCs are studied as compact nonlinear feature maps whose learning dynamics can be characterized through empirical NTK analysis and compared with closely matched classical baselines. Our objective is to identify differences in learning geometry rather than to assume a
universal quantum advantage.

\section{Methodology}

\subsection{Problem setting}
We study forecasting tasks on synthetic multivariate signals with controlled spectral complexity. Given an input sequence
\[
X \in \mathbb{R}^{L \times F},
\]
where $L$ is the sequence length and $F$ is the number of features, the goal is to predict the next value of a target variable at time step $L+1$. Our analysis focuses on comparing how a compact hybrid quantum model and a classical baseline  with comparable structure learn under increasingly challenging spectral regimes.

\subsection{Models under comparison}

\subsubsection{Hybrid quantum model}

The hybrid quantum model consists of a classical input projection, a variational quantum circuit (VQC), and a linear readout head. The input sequence is first projected to a latent dimension equal to the number of qubits. Only the last projected time step is used by the quantum block
and is scaled as:
\begin{equation}
    x_q = \pi\tanh\left(\widetilde{X}_{L,:}\right)
    \in \mathbb{R}^{Q},
\end{equation}
where $Q$ denotes the number of qubits.

Before data encoding, each qubit is prepared in an equal superposition,
yielding the initial state
\begin{equation}
    |\Psi_0\rangle
    =
    H^{\otimes Q}|0\rangle^{\otimes Q}
    =
    |+\rangle^{\otimes Q}.
\end{equation}
This preparation ensures that the subsequent $R_Z$ encoding introduces
input-dependent relative phases from the first encoding operation.

The VQC contains $n_{\mathrm{layers}}$ data re-uploading layers,
\begin{equation}
    U^{(l)}(x_q)
    =
    U_{\mathrm{ent}}
    \left[
        \bigotimes_{i=1}^{Q}
        \operatorname{Rot}
        (\phi_{l,i},\vartheta_{l,i},\omega_{l,i})
    \right]
    \left[
        \bigotimes_{i=1}^{Q}
        R_Z(s_l x_{q,i})
    \right],
\end{equation}
where $s_l$ is a trainable re-uploading scale and
$U_{\mathrm{ent}}$ applies CNOT gates in a ring topology. The input
representation is re-encoded at every circuit layer. The final state is
\begin{equation}
    |\Psi(x_q)\rangle
    =
    U^{(n_{\mathrm{layers}})}(x_q)
    \cdots
    U^{(1)}(x_q)
    |\Psi_0\rangle.
\end{equation}
One Pauli-$Z$ expectation value is measured per qubit,
\begin{equation}
    z_i =
    \langle\Psi(x_q)|Z_i|\Psi(x_q)\rangle,
    \qquad i=1,\ldots,Q,
\end{equation}
and the resulting feature vector is passed to a linear head to produce
the scalar forecast.

\subsubsection{Classical baseline}
To make the comparison as controlled as possible, we designed a classical baseline with the same  structure of the hybrid quantum model. As in the quantum case, the input sequence is first projected to a latent dimension of size $Q$ and scaled with $\pi \tanh(\cdot)$. Again, only the last time step is used for prediction.

The reuploading strategy of the quantum model is introduced in the classical part as well. To this end, let $y_0 \in \mathbb{R}^{Q}$ denote the projected last time step. The classical latent state is initialized as $h^{(0)} = y_0$, and then updated over $n_{\text{layers}}$ stages according to
\[
h^{(l+1)} = W^{(l)} \left( \pi \tanh\left(h^{(l)} + y_0\right) \right),
\]
where each $W^{(l)}$ is a trainable linear mixing layer without bias. In this way, the original input representation $y_0$ is re-injected at every layer, providing a classical analogue of the repeated data encoding performed by the quantum circuit. The number of classical layers is matched to the number of quantum reuploading layers.

Finally, the resulting latent representation is passed through a linear layer to generate the prediction. This construction does not make the two models identical, but it provides a closer comparison by ensuring that both architectures repeatedly mix the same latent input across multiple layers before the final readout.
\subsection{Trainable parameters}
Both models use a latent dimension of eight and four trainable
re-uploading stages. The resulting hybrid quantum and classical models
contain 125 and 281 trainable parameters, respectively. Thus, the
comparison matches latent dimension and depth, but not parameter count;
the classical baseline has approximately $2.25\times$ more trainable
parameters.
\subsection{Synthetic datasets}

\subsubsection{Harmonic-mixture dataset}
Our first benchmark consists of multivariate signals built from a sinusoidal component and a higher-frequency harmonic. For each sample and feature, amplitudes and phases are drawn uniformly at random :
\[
A \sim \mathcal{U}(0.5, 1.5), \qquad \phi \sim \mathcal{U}(0, 2\pi).
\]
Given a base frequency $f$, the signal is defined as
\[
x(t) = A \sin(2\pi f t + \phi) + 0.5 A \sin(2\pi (3f)t + 1.5\phi) + \epsilon(t),
\]
where $\epsilon(t) \sim \mathcal{N}(0, \sigma^2), \qquad \sigma = 0.2.$
The time grid is
\[
t = \left\{0, \frac{1}{L+1}, \frac{2}{L+1}, \dots, \frac{L}{L+1}\right\},
\]
implemented as a uniform partition of $[0,1)$ with $L+1$ points. The model input consists of the first $L$ points of the signal across all features, while the target is the next-step value of the first feature: $X = x_{1:L}, \qquad y = x_{L+1}$.

\subsubsection{Chirp dataset}
To test whether the observed behavior persists beyond stationary harmonic signals, we also consider nonstationary chirp signals. For each feature, amplitudes and phases are sampled from the same distributions as above. The base  chirp is constructed from a linearly varying instantaneous frequency from $f_0$ to $f_1$, whose integrated phase is
\[
\Phi_{\text{base}}(t) = 2\pi \left( f_0 t + \frac{1}{2}(f_1-f_0)t^2 \right).
\]
A harmonic chirp component is added using a corresponding sweep from $3f_0$ to $3f_1$:
\[
\Phi_{\text{harm}}(t) = 2\pi \left( 3f_0 t + \frac{1}{2}(3f_1-3f_0)t^2 \right).
\]
The resulting signal is
\[
x(t) = A \sin(\Phi_{\text{base}}(t) + \phi)
+ w_h A \sin(\Phi_{\text{harm}}(t) + 1.5\phi) + \epsilon(t),
\]
where $w_h = 0.5$ is the harmonic weight and $\epsilon(t)$ is Gaussian noise with standard deviation $\sigma = 0.2$. As in the harmonic dataset, the target is the next-step value of the first feature.

\subsubsection{Experimental regimes}

For the harmonic-mixture benchmark, we varied the base frequency over
$f\in\{1,2,4,8,12,16\}$ to generate tasks with increasing oscillatory
complexity. For the chirp benchmark, we fixed the initial frequency to
$f_0=2$ and varied the final frequency $f_1=\{8,12,16\}$. Both benchmarks were evaluated with 200 and 50 training samples to examine the effect of data-limited regimes.

Both architectures use a latent dimension of eight and four re-uploading stages. The hybrid model therefore employs
eight qubits and four rotation--entanglement layers, while the classical baseline uses four corresponding mixing layers. 
Both models were trained using the Adam optimizer~\cite{adam} with mean squared error (MSE) loss. To reduce optimizer mismatch between architectures, learning rates were selected independently by minimizing the mean AULC over 20 random seeds across
$\{0.001,0.003,0.01,0.03,0.1,0.3\}$, yielding $0.1$ for the hybrid quantum model and $0.03$ for the classical baseline. For each condition, fixed validation and test sets of 500 and 1000 samples were shared across models and runs. The 20 random seeds vary only parameter initialization. Validation MSE was evaluated at every epoch, and the checkpoint with the lowest validation loss was selected independently for each model and seed. Test MSE was then evaluated at the selected checkpoint. Training loss, AULC, and NTK diagnostics are computed on the training set.

\subsection{Empirical NTK analysis}
To characterize differences in learning beyond final predictive accuracy, we analyze training dynamics through the empirical NTK. For each model and selected training epoch, we compute the kernel induced by the Jacobian of the model outputs with respect to trainable parameters on the training set. This gives a local representation of how samples are related by the current parameterization of the model.

Our goal is not to derive new NTK theory, but to use NTK-based diagnostics as a principled empirical framework for comparing the learning geometry of the quantum and classical models.

From the empirical kernel, we compute the following quantities.

\paragraph{Kernel-target alignment.}
We quantify how strongly the kernel supports the target direction through
\[
\lambda_{\mathrm{eff}}(K,y)=\frac{y^\top K y}{\mathrm{Tr}(K)+\varepsilon}.
\]
This metric measures how well the target projects onto the dominant directions of the kernel. Larger values indicate that the kernel is better aligned with the target, which is generally associated with faster early-stage learning. 

\paragraph{Top-1 spectral mass.}
We compute the fraction of spectral mass carried by the leading eigenvalue:
\[
m_1(K)=\frac{\lambda_{\max}}{\sum_j \lambda_j}.
\]
This metric captures how concentrated the kernel spectrum is around its dominant mode. A larger top-1 mass indicates that the kernel is dominated by a single principal direction, whereas smaller values reflect a more distributed spectral structure.

\paragraph{Kernel drift.}
To measure how much the kernel changes during training relative to initialization, we compute
\[
\mathrm{drift}(K_t,K_0)=\frac{\|K_t-K_0\|}{\|K_0\|}.
\]
This quantity measures the relative deformation of the kernel geometry during optimization. Larger values indicate that the model departs more strongly from its initial kernel, suggesting a greater degree of kernel reconfiguration during training. Smaller values indicate that the model remains closer to its initialization, which is consistent with a more stable or near-frozen kernel regime.

Experiments were implemented in PennyLane [16] using default.qubit, the JAX-JIT interface, backpropagation, and vectorized circuit evaluation, and executed on an RTX 3080 GPU with a Ryzen 7 5700X CPU.

\section{Results}\label{sec2}
In the results below, we first examine four training diagnostics:
kernel-target alignment, kernel drift, top-1 spectral mass, and training
loss. Together, these quantities characterize target alignment, kernel
reconfiguration, spectral concentration, and in-sample optimization.
Validation-selected checkpoints and held-out test performance are analyzed
separately. In all figures, curves and shaded regions show the mean
$\pm$ one standard deviation over 20 initialization seeds.

\begin{figure}[t]
    \centering
    \includegraphics[width=1.0\linewidth]{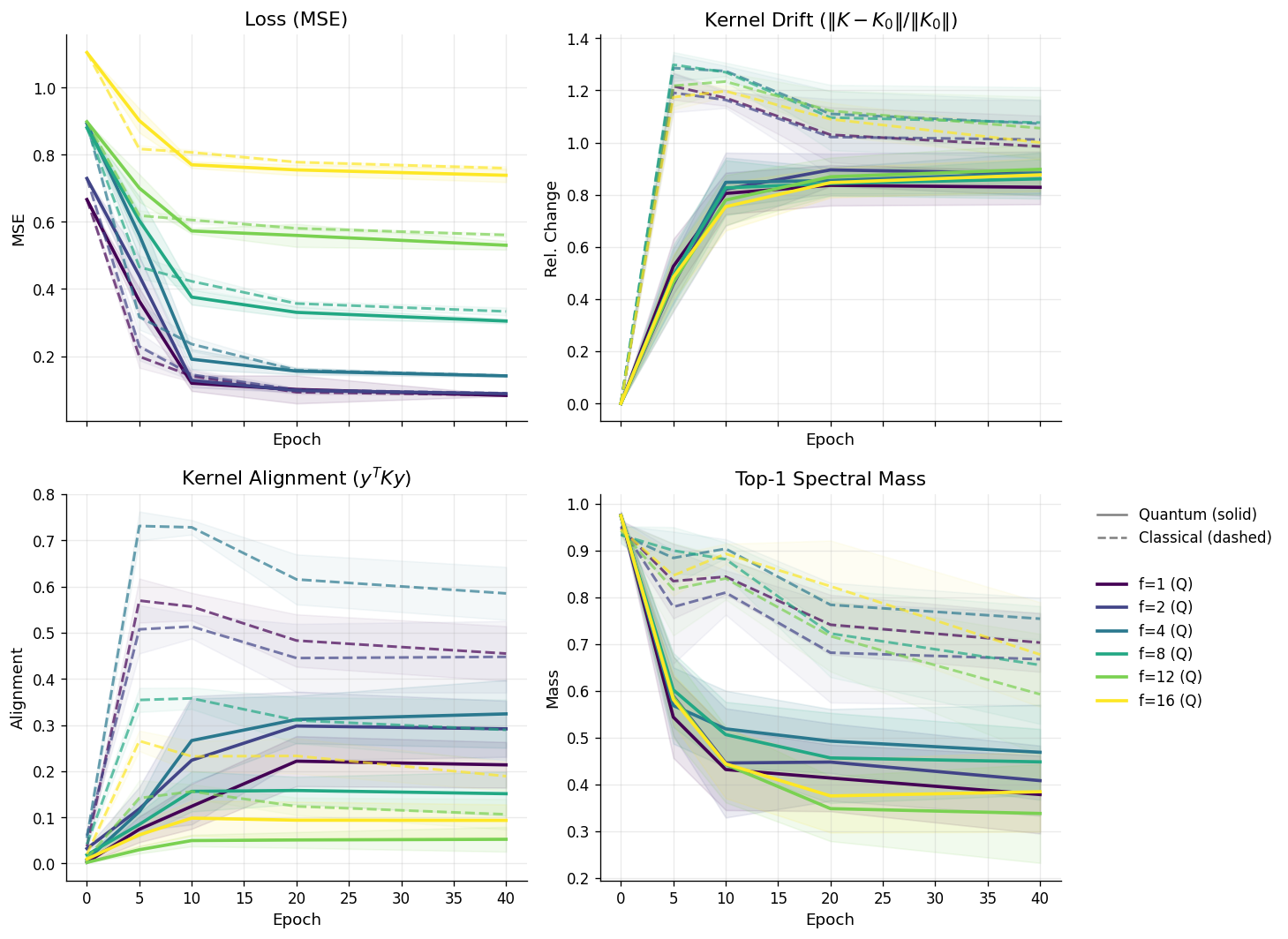}
    \caption{NTK Dynamics for the Harmonic-mixture dataset with 200 training samples.}
    \label{fig:ntk_harmonic_200}
\end{figure}

\begin{figure}[t]
    \centering
    \includegraphics[width=1.0\linewidth]{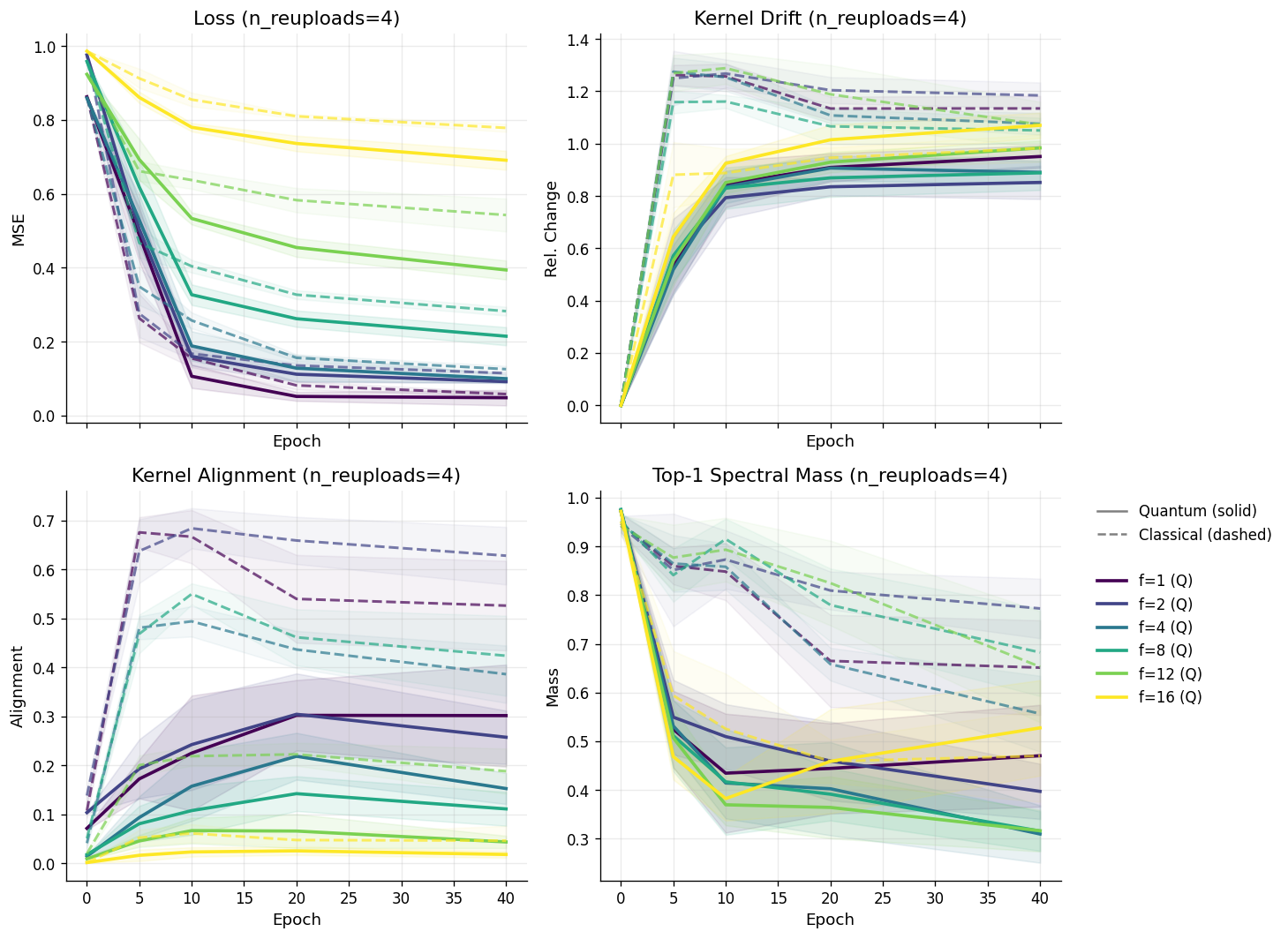}
    \caption{NTK Dynamics for the Harmonic-mixture dataset with 50 training samples.}
    \label{fig:ntk_harmonic_50}
\end{figure}

Figures~\ref{fig:ntk_harmonic_200} and
\ref{fig:ntk_harmonic_50} show the harmonic-mixture results with 200 and
50 training samples, respectively. Across both regimes, the classical
baseline exhibits stronger early kernel-target alignment and larger
kernel drift, whereas the hybrid model remains closer to its initial
kernel and develops a less concentrated spectrum. The training-loss
trajectories also differ across spectral regimes, with the separation
becoming more pronounced in the 50-sample setting. These results reveal
persistent differences in the optimization geometry whose relation
to validation convergence is examined below.

\begin{figure}[t]
    \centering
    \includegraphics[width=1.0\linewidth]{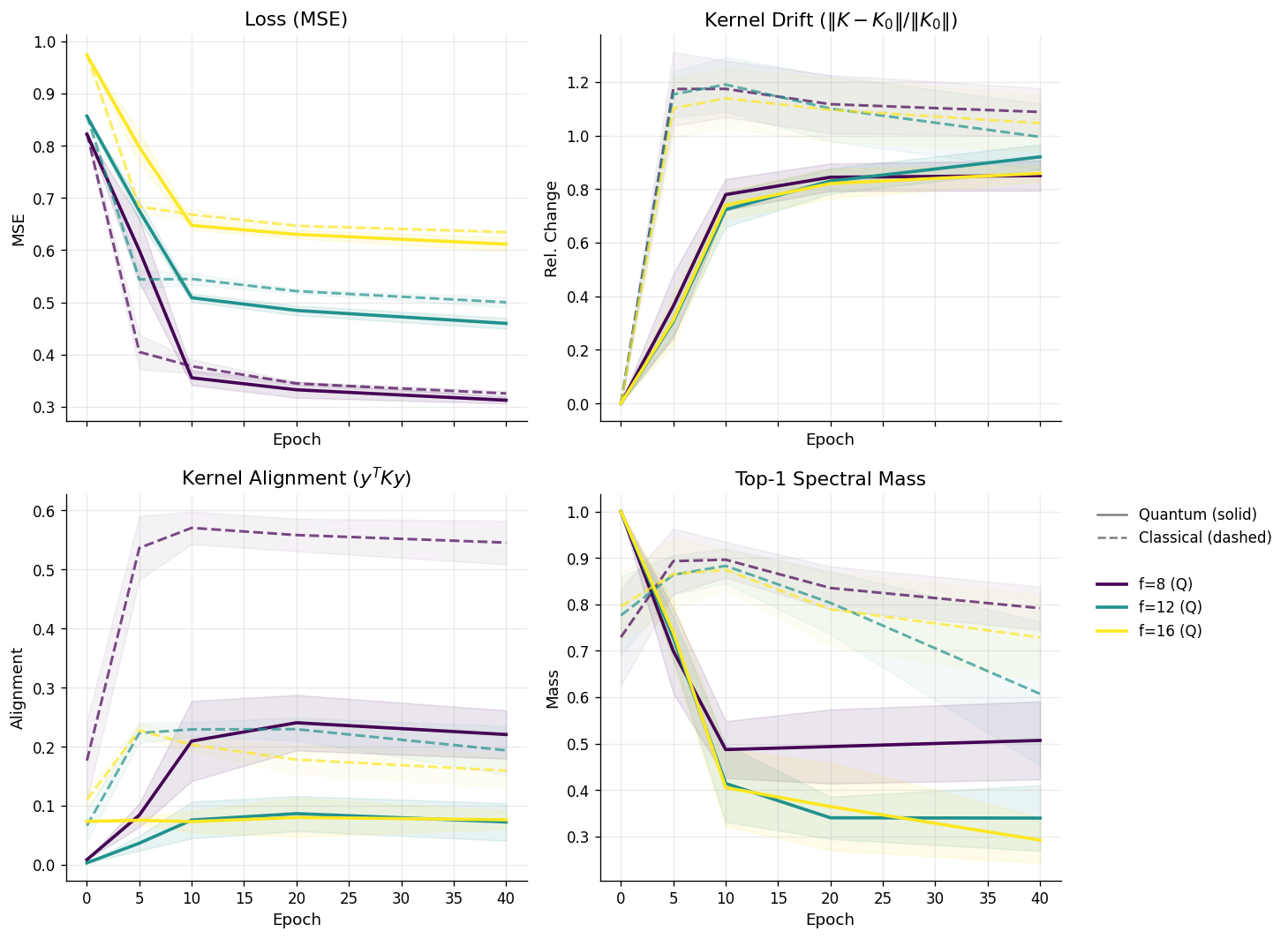}
    \caption{NTK Dynamics for the Chirp dataset with 200 training samples.}
    \label{fig:ntk_chirp_200}
\end{figure}

\begin{figure}[t]
    \centering
    \includegraphics[width=1.0\linewidth]{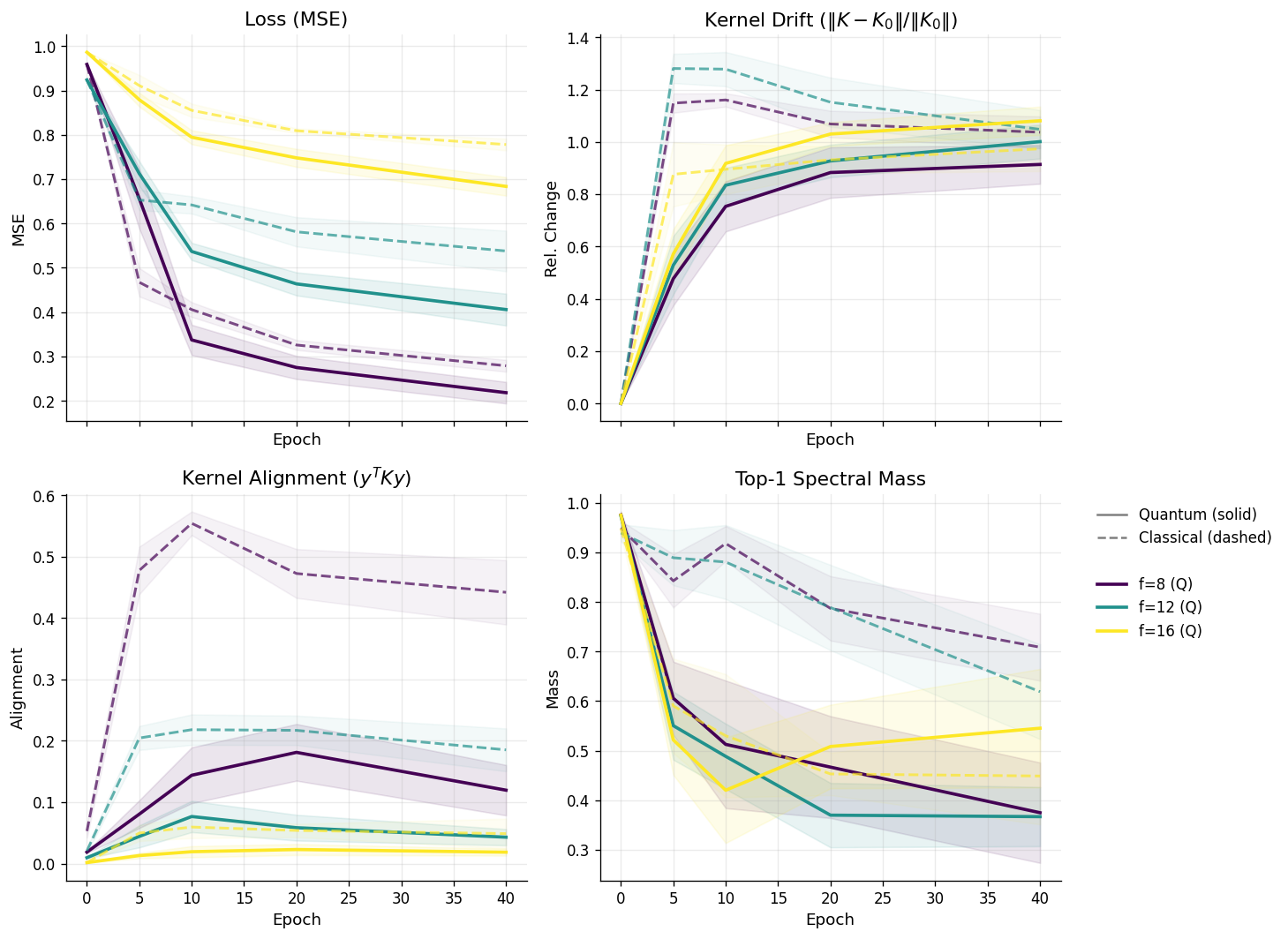}
    \caption{NTK Dynamics for the Chirp dataset with 50 training samples.}
    \label{fig:ntk_chirp_50}
\end{figure}

Figures~\ref{fig:ntk_chirp_200} and \ref{fig:ntk_chirp_50} extend the analysis to nonstationary chirp signals.
The classical baseline again develops stronger kernel-target alignment and abrupt early kernel drift, whereas the hybrid model evolves more gradually and generally maintains a less concentrated spectrum. The difference in training-loss trajectories is particularly visible with 50 samples, where the classical model degrades more strongly relative to
the 200-sample regime. Thus, the characteristic kernel-level differences persist beyond stationary harmonic signals and remain visible under limited training data.

Table~\ref{tab:frequency_dynamics} examines the frequency dependence of the learning dynamics in the 50-sample harmonic-mixture regime. The hybrid model reaches its validation-selected checkpoint earlier at five of the six evaluated frequencies, with the largest epoch reduction at $f=12$ ($58.73\%$). Significant paired differences in checkpoint epoch are observed for $f=1,2,4,8,$ and $12$. However, this behavior does not directly track training AULC. At lower frequencies, earlier validation selection occurs despite a higher cumulative training loss, whereas at $f=12$ both metrics favor the hybrid model. At $f=16$, the AULC advantage persists although the classical model reaches its selected checkpoint slightly earlier. Taken together, these results reveal a decoupling between cumulative training efficiency and validation-selected convergence: lower training AULC does not
systematically imply earlier attainment of the best validation solution.

\begin{table}[t]
\centering
\caption{Frequency-dependent learning dynamics for the 50-sample harmonic-mixture benchmark. Positive epoch reduction indicates earlier attainment of the validation-selected checkpoint by the hybrid model; positive AULC gain indicates lower training AULC. $p_E$ and $p_A$ correspond to paired Wilcoxon signed-rank tests over 20 initialization seeds.}
\label{tab:frequency_dynamics}
\begin{tabular}{ccccc}
\toprule
Freq & Epoch reduction(\%) & $p_E$ &Train AULC gain(\%) & $p_{A}$ \\
\midrule
1 & 11.87 & 0.0048& -9.44 &  0.0012 \\
2 & 9.20 & 0.0185 & -12.40 & $<$0.0001 \\
4 & 26.80 & 0.0197&  -5.35 & 0.0353 \\
8 & 24.36 & 0.00316 &-0.35 & 0.8469 \\
12 & 58.73 & 0.0009 & 8.84 & $<$0.0001 \\
16 & -6.06 & 0.0990 & 5.72 & $<$0.0001 \\
\bottomrule
\end{tabular}
\end{table}
Table~\ref{tab:benchmark_summary} summarizes validation-selected
convergence and held-out performance across benchmarks. The hybrid model
reaches its selected checkpoint earlier in five of six harmonic
conditions and in two of three evaluated chirp conditions. In contrast, test gains are small and do not reveal a consistent held-out advantage, with aggregated test differences remaining within $3.6\%$. This parity is obtained despite the hybrid model using approximately $55\%$ fewer trainable parameters. Thus, comparable held-out performance is reached through distinct optimization trajectories, with the hybrid model more frequently
attaining its validation-selected solution earlier.
\begin{table}[t]
\centering
\caption{Validation-selected convergence and held-out performance across
forecasting benchmarks. Values are averaged across evaluated frequencies
and 20 initialization seeds. Positive test gain indicates lower test MSE
for the hybrid model; epoch reduction indicates earlier attainment of the
validation-selected checkpoint. The last column indicates in how many frequency the quantum model achieved the checkpoint earlier}
\label{tab:benchmark_summary}
\begin{tabular}{lccccc}
\toprule
Dataset & $N$ & Test gain ($\%$) & Epoch reduction &  Q earlier \\
\midrule
Harmonic & 200 & -2.21 & 4.72 & 5/6 \\
Harmonic & 50  & -1.07 & 5.43 & 5/6 \\
Chirp    & 200 & 1.40 & 11.53 & 3/3 \\
Chirp    & 50  & -3.58 & 3.12 & 2/3 \\
\bottomrule
\end{tabular}
\end{table}

To assess whether the observed performance differences can be explained solely by the spectral structure introduced by quantum angle encoding, we considered an additional classical baseline augmented with explicit Fourier features. The model preserves the input projection, data re-uploading structure, number of mixing layers, and linear readout of the original classical baseline. However, before each mixing layer, the latent representation is expanded using sinusoidal features $\phi(z)=[z, \{sin(bz),cos(bz)\}_
{b\in B}]$ over a fixed set of frequency bands $B=(1,2,4,8,16)$, together with the raw latent features. The resulting feature vector is normalized by the square root of its dimension and projected back to the $Q$-dimensional latent space. This baseline provides the classical model with an explicit periodic representation while retaining the overall re-uploading architecture used in the original comparison. 



The Fourier-augmented classical baseline does not eliminate the optimization gap observed on the harmonic-mixture benchmark.
Across the complete frequency sweep, the hybrid model achieves
$7.63$--$20.31\%$ lower final training loss and $18.70$--$45.90\%$
lower training AULC. The final-loss differences are significant at all
six frequencies ($p\leq0.0194$), while the AULC differences satisfy
$p<0.0001$ throughout. These results indicate that explicit sinusoidal
features alone are insufficient to explain the observed differences in
training dynamics.

\begin{figure}[t]
    \centering
    \includegraphics[width=1.0\linewidth]{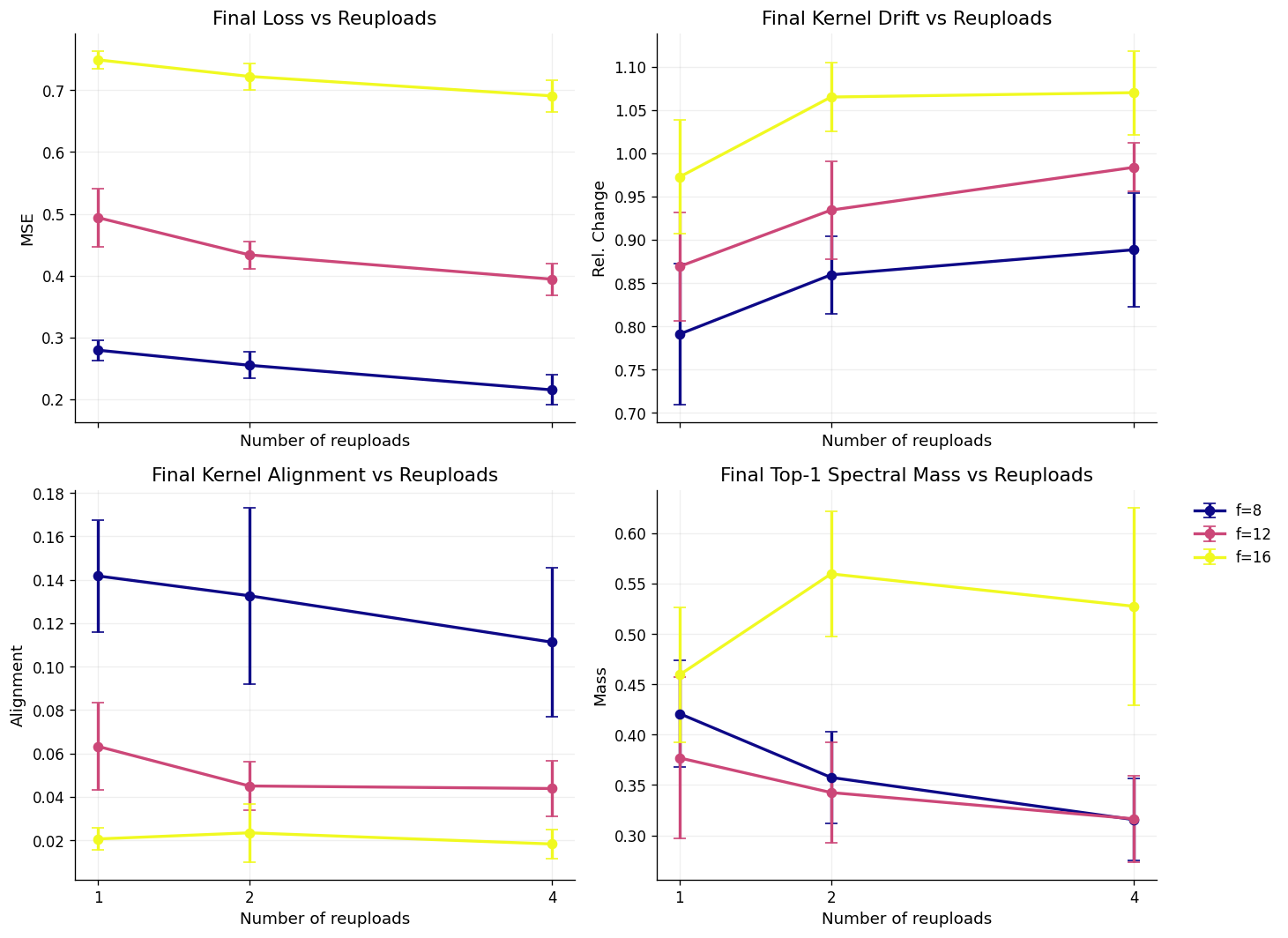}
    \caption{Effect of the number of data re-uploading operations on final
loss and empirical NTK metrics for the harmonic-mixture benchmark with
50 training samples. Circuit depth is fixed to four
rotation--entanglement layers.}
    \label{fig:reuploading}
\end{figure}

To isolate the contribution of repeated data encoding, we vary the
number of re-uploading operations while keeping the four trainable
rotation--entanglement layers fixed. Figure~\ref{fig:reuploading}
shows that increasing the number of re-uploads consistently reduces
final loss across the three representative frequency regimes. This
improvement is accompanied by larger kernel drift and, particularly for
$f=8$ and $f=12$, lower top-1 spectral mass. Kernel-target alignment
does not increase with additional re-uploading and instead remains
stable or decreases. These results show that repeated encoding affects both predictive performance and the learned kernel geometry. They also indicate that stronger global alignment or smaller kernel drift, when considered individually, are not sufficient indicators of lower final error.

\subsection{Discussion}
The central result of this study is not a systematic difference in endpoint generalization, but the observation that comparable held-out performance can arise from substantially different learning geometries. The 125-parameter hybrid model remains within a few percent of the 281-parameter classical baseline across all evaluated regimes, while more frequently reaching its validation-selected solution in fewer optimization epochs. At the same time, the architectures exhibit persistent differences in kernel-target alignment, kernel drift, and spectral concentration. Across the experimental settings, the empirical NTK diagnostics reveal a recurring distinction between the learning dynamics of the hybrid quantum and classical models. The classical baseline generally exhibits stronger early kernel-target alignment and faster initial loss reduction, whereas the hybrid model tends to remain closer to its initial kernel and develop a less concentrated spectral structure. These patterns remain qualitatively visible for both stationary harmonic-mixture and nonstationary chirp signals.

Importantly, the hybrid model preserves broadly comparable held-out performance despite using substantially fewer trainable parameters and following a systematically different optimization trajectory. The frequency  analysis reveals a decoupling between cumulative training efficiency and validation-selected convergence: the hybrid model frequently reaches its best validation checkpoint earlier even in regimes where its training AULC is higher. Moreover, stronger early kernel-target alignment in the classical model is not systematically associated with earlier validation convergence. Together, these observations indicate that neither cumulative training loss nor a single NTK diagnostic provides a monotonic metric for how rapidly the best validation solution emerges.

The Fourier-feature comparison and re-uploading ablation further support this interpretation. The persistence of the optimization gap under explicit Fourier augmentation shows that the observed learning dynamics are not reproduced by periodic feature expansion alone. Moreover, increasing the number of quantum data re-uploads reduces final training
loss despite increasing kernel drift and without strengthening kernel-target alignment. This second form of decoupling reinforces the observation that individual kernel diagnostics cannot be interpreted as monotonic indicators of optimization quality. Rather, learning behavior appears to emerge from the joint evolution of kernel alignment, spectral organization, and kernel reconfiguration.

Instead, the interaction between repeated data encoding and trainable quantum transformations appears to contribute to the resulting kernel geometry. However, the present experiments do not isolate a uniquely quantum mechanism, as the hybrid architecture combines angle encoding, trainable rotations, and entangling operations. Dedicated ablations of
these components remain necessary.

The observed results also cannot be explained by a larger quantum model: the hybrid and classical architectures contain 125 and 281 trainable parameters, respectively. Nevertheless, parameter count is not a direct measure of equivalent representational capacity across different architectures; the comparison should therefore be interpreted as structurally aligned rather than parameter matched.

Overall, empirical NTK analysis provides a useful perspective for characterizing hybrid quantum models beyond endpoint accuracy. Our results identify reproducible differences in optimization geometry and convergence epoch, despite presenting comparable testing metrics. Because the quantum experiments use exact expectation values on a noiseless state-vector simulator, whether these kernel-level patterns persist under finite-shot and hardware noise remains an open question.
Future work should investigate their architectural origin and robustness under realistic quantum noise and more diverse classical baselines.

\section{Conclusion}\label{sec13}

In this work, empirical NTK diagnostics reveal reproducible differences between the learning geometries of a compact hybrid quantum forecasting model and a structurally aligned classical baseline. The classical model generally exhibits stronger early kernel-target alignment and faster
initial training-loss reduction, whereas the hybrid model tends to develop a less concentrated kernel spectrum and undergo smaller kernel drift.

These differences are accompanied by a non-monotonic relationship between optimization efficiency and validation-selected convergence. Thus, earlier validation convergence is not systematically predicted by training AULC or stronger kernel-target alignment.

A Fourier-augmented classical baseline does not eliminate the observed in-sample optimization gap, while the re-uploading ablation shows that lower training loss can emerge despite increased kernel drift and without
stronger alignment. Together, these findings show that individual NTK diagnostics do not provide monotonic relations for learning behavior across the studied architectures. Rather than establishing a general quantum advantage, our results identify systematically different optimization
trajectories that reach comparable held-out performance and motivate a more joint interpretation of kernel alignment, spectral organization, and kernel evolution.

\bibliography{references}

\end{document}